\documentclass[11pt,a4paper]{article}
\usepackage[hyperref]{eacl2021}
\usepackage{times}
\usepackage{latexsym}
\usepackage{multirow}
\usepackage{graphicx}

\usepackage{svg}

\usepackage{tabularx}
\newcolumntype{R}[1]{>{\hsize=#1\hsize\raggedleft\arraybackslash}X}%
\newcolumntype{L}[1]{>{\hsize=#1\hsize\raggedright\arraybackslash}X}%
\newcolumntype{C}[1]{>{\hsize=#1\hsize\centering\arraybackslash}X}%

\usepackage{microtype}
\usepackage{adjustbox}

\aclfinalcopy 

\title{From Toxicity in Online Comments to Incivility in American News: Proceed with Caution}

\author{
  Anushree Hede \qquad Oshin Agarwal \qquad Linda Lu \qquad Diana C. Mutz \qquad  Ani Nenkova \\
  University of Pennsylvania \\ 
  {\tt \{anuhede,oagarwal,lulinda,nenkova\}@seas.upenn.edu, mutz@upenn.edu} }

\date{}

\begin{document}
\maketitle
\begin{abstract}
The ability to quantify incivility online, in news and in congressional debates, is of great interest to political scientists. Computational tools for detecting online incivility for English are now fairly accessible and potentially could be applied more broadly. We test the Jigsaw Perspective API for its ability to detect the degree of incivility on a corpus that we developed, consisting of manual annotations of civility in American news. We demonstrate that toxicity models, as exemplified by Perspective, are inadequate for the analysis of incivility in news. We carry out error analysis that points to the need to develop methods to remove spurious correlations between words often mentioned in the news, especially identity descriptors and incivility. Without such improvements, applying Perspective or similar models on news is likely to lead to wrong conclusions, that are not aligned with the human perception of incivility. 
\end{abstract}

\section{Introduction}

Surveys of public opinion report that most Americans think that the tone and nature of political debate in this country have become more negative and less respectful and that the heated rhetoric by politicians raises the risk for violence \cite{pew19}.   
These observations motivate the need to study (in)civility in political discourse in all spheres of interaction, including online \cite{ziegele2018dynamics,jaidka2019brevity}, in congressional debates \cite{uslaner2000senate} and as presented in news \cite{meltzer2015journalistic,rowe2015civility}. Accurate automated means for coding incivility could facilitate this research, and political scientists have already turned to using off-the-shelf computational tools for studying civility \cite{frimer2018montagu,jaidka2019brevity,theocharis2020dynamics}.

Computational tools however, have been developed for different purposes, focusing on detecting language in online forums that violate community norms.
The goal of these applications is to support human moderators by promptly focusing their attention on likely problematic posts. When studying civility in political discourse, it is primarily of interest to characterize the overall civility of interactions in a given source (i.e., news programs) or domain (i.e., congressional debates), as an average over a period of interest. Applying off-the-shelf tools for toxicity detection is appealingly convenient, but such use has not been validated for any domain, while uses in support of moderation efforts have been validated only for online comments.  

We examine the feasibility of quantifying incivility in the news via the Jigsaw Perspective API, which has been trained on over a million online comments rated for toxicity and deployed in several scenarios to support moderator effort online\footnote{\url{https://www.perspectiveapi.com/}}. We collect human judgments of the (in)civility in one month worth of three American news programs.
We show that while people perceive significant differences between the three programs, Perspective cannot reliably distinguish between the levels of incivility as manifested in these news sources.

We then turn to diagnose the reasons for Perspective's failure. Incivility is more subtle and nuanced than toxicity, which includes identity slurs, profanity, and threats of violence along other unacceptable incivility.  In the range of civil to borderline civil human judgments, Perspective gives noisy predictions that are not indicative of the differences in civility perceived by people. This finding alone suggests that averaging Perspective scores to characterize a source is unlikely to yield meaningful results. To pinpoint some of the sources of the noise in predictions, we characterize individual words as likely triggers of errors in Perspective or sub-error triggers that lead to over-prediction of toxicity.  

We discover notable anomalies, where words quite typical in neutral news reporting are confounded with incivility in the news domain.
We also discover that the mention of many identities, such as Black, gay, Muslim, feminist, etc., triggers high incivility predictions. This occurs despite the fact that Perspective has been modified specifically to minimize such associations \cite{DBLP:conf/aies/DixonLSTV18}. Our findings echo results from gender debiasing of word representations, where bias is removed as measured by a fixed definition but remains present when probed differently \cite{gonen-goldberg-2019-lipstick-pig}. This common error---treating the mention of identity as evidence for incivility---is problematic when the goal is to analyze American political discourse, which is very much marked by us-vs-them identity framing of discussions. 

These findings will serve as a basis for future work in debiasing systems for incivility prediction, while the dataset of incivility in American news will support computational work on this new task. 

Our work has implications for researchers of language technology and political science alike. For those developing automated methods for quantifying incivility, we pinpoint two aspects that require improvement in future work: detecting triggers of civility overprediction and devising methods to mitigate the errors in prediction. We propose an approach for a data-driven detection of error triggers; devising mitigation approaches remain an open problem. For those seeking to contrast civility in different sources, we provide compelling evidence that state-of-the-art automated tools are not appropriate for this task.  
The data and (in)civility ratings would be of use to both groups as test data for future models for civility prediction\footnote{Available at \url{https://github.com/anushreehede/incivility_in_news}}.

\section{Related Work}

\subsection{Datasets for Incivility Detection}

Incivility detection is a well-established task, though it is not well-standardized, with the degree and type of incivility varying across datasets.

Hate speech, defined as speech that targets social groups with the intent to cause harm, is arguably the most widely studied form of incivility detection, largely due to the practical need to moderate online discussions. Many Twitter datasets have been collected, of racist and sexist tweets \cite{waseem-hovy-2016-hateful}, of hateful and offensive tweets \cite{davidson2017automated}, and of hateful, abusive, and spam tweets \cite{founta2018large}. Another category of incivility detection that more closely aligns with our work is toxicity prediction. \citet{wulczyn2017ex} collected a dataset for toxicity identification in online comments on Wikipedia talk pages. They defined toxicity as comments that are rude, disrespectful, or otherwise likely to make someone leave a discussion. All these datasets are built for social media platforms using either online comments or tweets. We work with American news.
 
To verify if Perspective can reproduce human judgments of civility in this domain, we collect a corpus of news segments annotated for civility. 

\subsection{Bias in Incivility Detection}

Models trained on the datasets described above associate the presence of certain descriptors of people with incivility  \citet{wulczyn2017ex}. 
This bias can be explained by the distribution of words in incivility datasets and the fact that systems are not capable of using context to disambiguate between civil and uncivil uses of the word, instead associating the word with the dominant usage in the training data. 

To mitigate this bias, Jigsaw's Perspective API was updated, and model cards \cite{mitchell2019model} were released for the system to show how well the system is able to predict toxicity when certain identity words are mentioned in a text. Simple templates such as ``I am $<$IDENTITY$>$'' were used to measure the toxicity associated with identity words. More recently, many more incorrect associations with toxicity were discovered. \citet{prabhakaran-etal-2019-perturbation} found that Perspective returned a higher toxicity score when certain names are mentioned, and \citet{hutchinson-etal-2020-social} found that this was also the case for words/phrases representing disability. ``I am a blind person'' had a significantly higher toxicity score than ``I am a tall person''. 
We show that when measured with different templates, the bias that was mitigated in Perspective still manifests. Further, we propose a way to establish a reference set of words and then find words associated with markedly higher toxicity than the reference. This approach reveals a larger set of words which do not lead to errors but trigger uncommonly elevated predictions of toxicity in the lower ranges of the toxicity scale.

\citet{waseem-hovy-2016-hateful} found that the most common words in sexist and racist tweets in their corpus are ``not, sexist, \#mkr, women, kat'' and ``islam, muslims, muslim, not, mohammed''. 
Any system trained on this dataset would likely learn these correlations with the frequent words as well, and computational methods to prevent this are yet to be developed. 

Studies have shown that African American English is more likely to be labeled as toxic by both annotators and systems due to differences in dialect \cite{sap-etal-2019-risk, xia-etal-2020-demoting, davidson-etal-2019-racial}. Our work does not involve AAE but does reveal how words quite common in news reporting can be erroneously associated with incivility.

\section{Data Collection for Incivility in News}

We study the following American programs: PBS NewsHour, MSNBC's The Rachel Maddow Show and Hannity from FOX News.
For brevity, we use the network to refer to each source: PBS, MSNBC, and FOX, in the following discussions. These sources are roughly representative of the political spectrum in American politics, with NewsHour being Left-Center
and MSNBC and FOX with strong left and right bias respectively. These are generally one-hour shows, with about 45 minutes of content when commercial breaks are excluded.

The transcripts, which provide speaker labels and turns, are from February 2019. We take only the days when all three shows aired, and we had transcripts, for a total of 51 transcripts.

We analyze the programs on two levels with the help of paid research assistants paid \$15 per hour. All of them are undergraduate students in non-technical majors at the University of Pennsylvania. In Pass I, we ask a single rater to read through a transcript and identify speaker turns that appear particularly uncivil or notably civil. We characterize the programs according to the number of uncivil segments identified in each transcript.\footnote{For this first analysis of incivility, we pre-selected uncivil segments, to ensure that they are well represented in the data. Incivility is rare, so a random sample of snippets will contain considerably fewer clearly uncivil snippets. We are currently augmenting the dataset with data from later months, with segments for annotation drawn from randomly selected time-stamps, giving a more representative sample of the typical content that appears in the show.} 

After that, in Pass II, a research assistant chose a larger snippet of about 200 words that includes the selection identified as notably uncivil (and respectively civil), providing more context for the initially selected speaker turns. We also selected several snippets of similar length at random but not overlapping with the civil and uncivil snippets. Each of these snippets, 219 in total, was then rated for perceived civility by two raters, neither of whom participated in the initial selection of content. The raters did not know why a snippet that they were rating was selected. We choose to annotate such snippets, corresponding to about a minute of show content, to ensure sufficient context is available to make meaningful judgments about the overall tone of presentation. Table \ref{tab:snippet_corpus_stats} gives an overview of number of Uncivil, Civil and Random speaker turns around which longer snippets were selected for fine-grained annotation of incivility. The largest number of civil turns was found in PBS and most uncivil turns were identified in FOX.

\begin{table}[t]
\centering 
\footnotesize
\setlength{\tabcolsep}{2pt}
\begin{tabularx}{\linewidth}{L{0.25}|R{0.1}R{0.1}R{0.1}|R{0.2}|R{0.2}|R{0.2}}
\textbf{Show} & \multicolumn{1}{c}{\bf U} & \multicolumn{1}{c}{\bf C} & \multicolumn{1}{c|}{\bf R} & \multicolumn{1}{c|}{\bf Overall} & \multicolumn{1}{c|}{\bf Avg Len} & \multicolumn{1}{c}{\bf Vocab} \\ 
\hline
FOX & 81 & 12 & 23 & 116 & 201.0 & 3960 \\
MSNBC & 13 & 23 & 8 & 44 & 209.6 & 1763 \\
PBS & 11 & 31 & 17 & 59 & 216.4 & 2627 \\ 
\hline
Overall & 105 & 66 & 48 & 219 & 206.9 & 5962  
\end{tabularx}
\caption{Number of Pass II snippets, separated by show and Pass I class: U(ncivil), C(ivil) and R(andom), along with average length of the snippets in words and the size of the vocabulary (unique words).}
\label{tab:snippet_corpus_stats}
\end{table}

\begin{figure}[t]
	\centering
	\includegraphics[scale=0.6
,width=\linewidth]{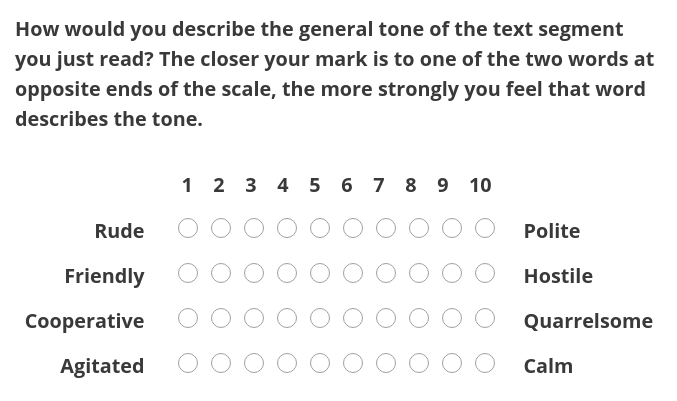}
	\caption{Annotation interface}
	\label{fig:interface}
\end{figure}

Snippets were presented for annotation in batches of 15. Each batch had a mix of snippets from each of the programs, displayed in random order. The annotators read each snippet and rated it on a 10 point scale in four dimensions used in prior work to assess civility in news \cite{mutz2005new}: Polite/Rude, Friendly/Hostile, Cooperative/Quarrelsome and Calm/Agitated. The dimensions appeared in random order and alternated which end of the dimension appeared on the left and which on the right, prompting raters to be more thoughtful in their selection and making it difficult to simply select the same value for all. A screenshot of the interface is shown in Figure \ref{fig:interface}. 

The composite civility score is then obtained by first reversing the ratings to account for the alternations of the ends for each dimension, such that all ratings result in small ratings for the civil ends of the scale (Polite, Friendly, Cooperative, Calm) and high values for the uncivil ends of the scales. The four scores were averaged for each annotator. Finally, the scores for the two annotators are averaged to obtain a civility score for each snippet, ranging between 1 and 10, where 1 is the most civil and 10 is the most uncivil possible.

\section{Annotator Agreement}
Here we briefly discuss the annotator agreement on the perceived incivility. We characterize the agreement on the transcript level in terms of the civil and uncivil speaker turns flagged in Pass I and on the text snippet level in terms of correlations and absolute difference in the scores assigned by a pair of raters in Pass II.

\begin{figure}[t]
	\centering
	\includegraphics[scale=0.6
,width=\linewidth]{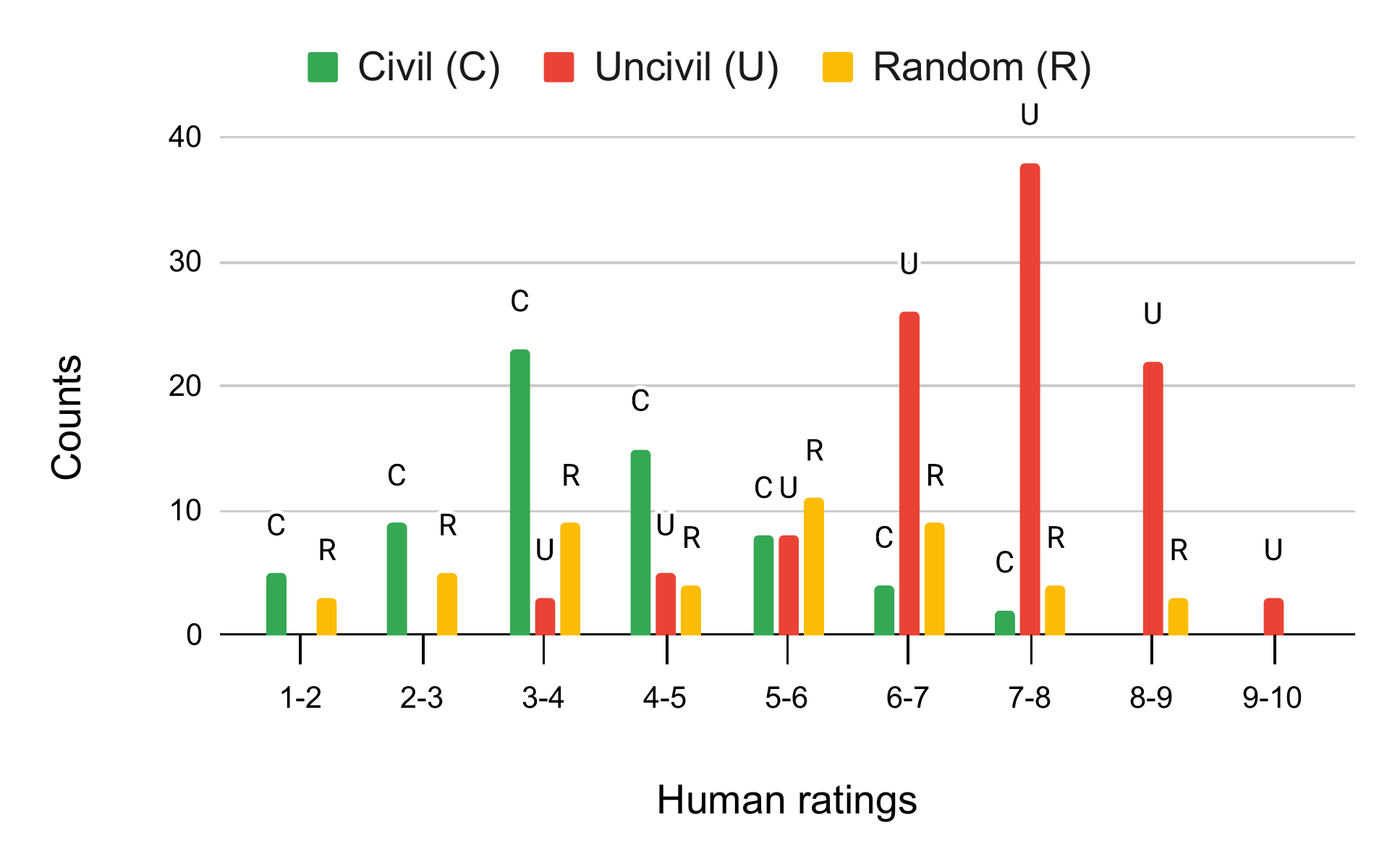}
	\caption{Distribution of snippet civility ratings (obtained in Pass II) by rationale type (obtained in Pass I). The snippets deemed to contain uncivil during Pass I are consistently rated as highly uncivil in Pass II.}
	\label{fig:score_distribution}
\end{figure}

Pass I selection of turns was made by one person, but we are still able to glean some insights about the validity of their selection from analyzing the ratings of Pass II snippets. The Pass I selection can be construed as a justification for the score assigned in Pass II, similar to other tasks in which a rationale for a prediction has to be provided \cite{deyoung-etal-2020-eraser}. Figure \ref{fig:score_distribution} shows the distribution of scores for the 200-word snippets that were selected around the initial 40-50-word speaker turns deemed notably civil, uncivil or that were randomly chosen. The distribution is as expected, with segments including uncivil turns almost uniformly rated as uncivil, with civility score greater than 5. According to our scale, a score of 5 would correspond to borderline civil on all dimensions or highly uncivil on at least one.

Only three of the 105 snippets selected around an uncivil turn got scores a bit under 5: the remaining snippets including a rationale for incivility, were rated as uncivil with high consistency by the independent raters in Pass II.

In Pass II, each pair of annotators rated a total of 37 or 36 segments. Correlations between the composite ratings are overall high, ranging from 0.86 to 0.65 per pair. For only one of the six pairs, the correlation is below 0.75. The absolute difference in composite incivility scores by the two independent annotators is about 1 point.

Overall, the consistency between initial selection from the transcript and the independent ratings of civility, as well as the correlation between civility ratings of two raters for the snippets, are very good.

\section{Predicting Incivility in American News}

With the corpus of real-valued civility in the news, we can now compare sources according to the perception of people with those provided by Perspective. We perform the analysis on the transcript level and the snippet level. On the transcript level, we use the number of uncivil utterances marked in Pass I as the indicator of incivility. For the snippet level, we use the average composite civility ratings by the raters in Pass II. For automatic analysis, we obtain Perspective scores for each speaker turn as marked in the transcript and count the number of turns predicted as having toxicity 0.5 or greater. For snippets, we obtain Perspective scores directly, combining multiple speaker turns. 

\begin{figure}[t]
	\centering
	\includegraphics[scale=0.48,trim={1.1cm 1cm 0.3cm 0cm},clip]{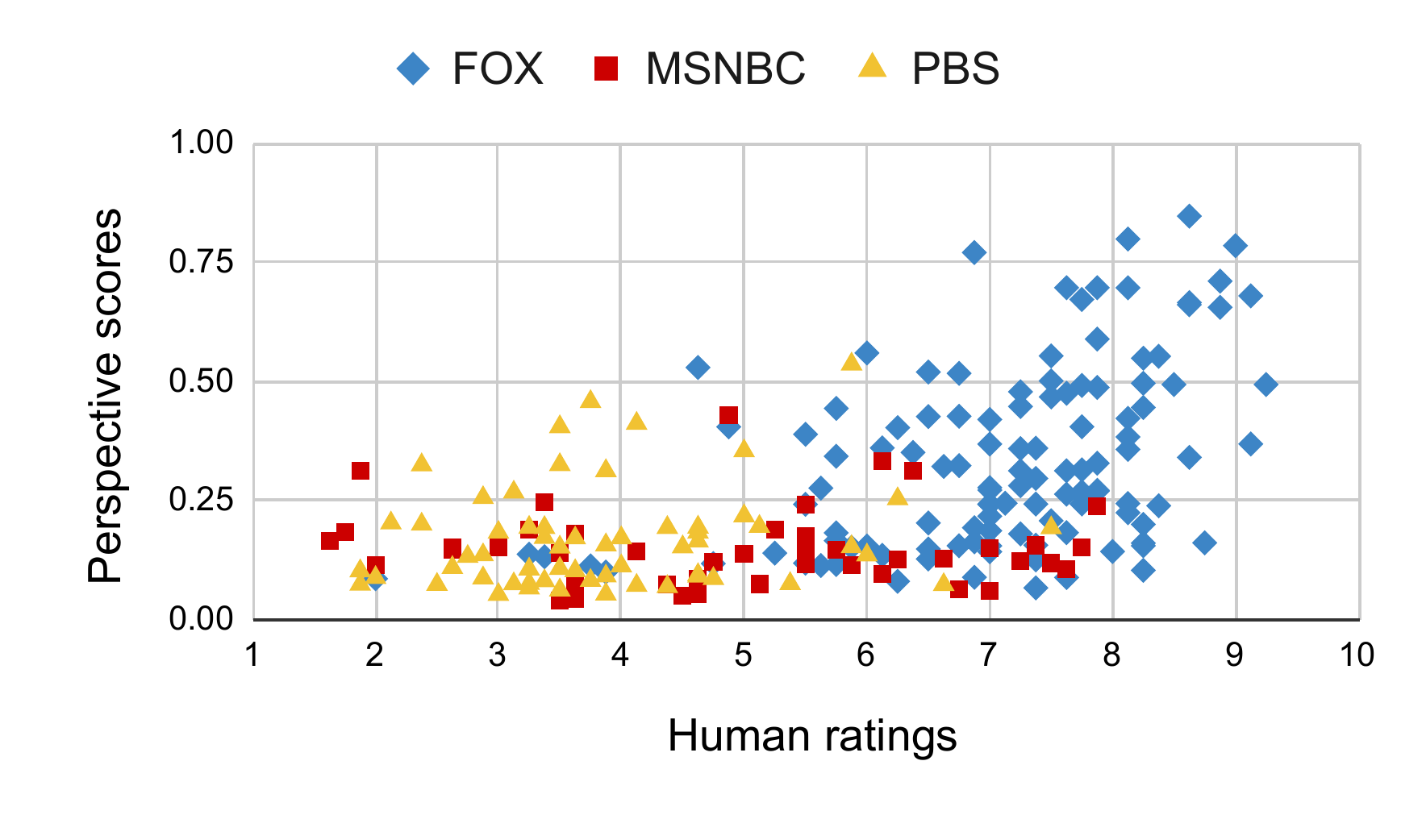}
	\caption{Scatterplot of human and Perspective scores}
	\label{fig:plot}
\end{figure}

Figure \ref{fig:plot} gives the scatter plot of civility scores per segment, from raters and Perspective. The plot reveals that Perspective is not sensitive enough to detect any differences in levels of incivility for human rating lower than six. For the higher levels of incivility, Perspective scores also increase and have better differentiation. However, the snippets from MSNBC rated as uncivil by people receive low scores. We verified that these segments are indeed uncivil, but in a sarcastic, indirect way. Portions of the turns can be seen in Table \ref{tab:maddow_eg}. 

\begin{table*}[t]
\centering
\scriptsize
\setlength{\tabcolsep}{4pt}
\begin{tabularx}{\linewidth}{L{1.0}|C{0.1}|C{0.1}}
\multicolumn{1}{c|}{\bf \small Snippets from MSNBC} & {\bf \small Human} & {\bf \small Perspective} \\ \hline
... They say that he is also going to declare the national emergency, but with this presidency, honestly, is that actually going to happen? We don't know. The White House says it's going to happen, but in this presidency, is anything really done and dusted before Fox \& Friends says it's done and dusted? & \multirow{2}{*}{\small 7.50} & \multirow{2}{*}{\small 0.12} \\ 
\hline
... Well, he better take a few moments, I know he doesn't read, but perhaps someone could read Article I Section 7 and 8 of the Constitution. The power of appropriation lies with the Congress, not with the president. If he were trying to do this, he is basically establishing an imperial president, his majesty... & \multirow{2}{*}{\small 7.75} & \multirow{2}{*}{\small 0.15} \\
\hline
proposing fireworks for the Fourth of July. Even in D.C., it's a bold idea from the president today. Presumably this will be followed by an executive order proclaiming that from here on out, we're going to start a whole new calendar year every year on the first day of January. What? Also, he's going to declare that we're going to start a new American pastime, a ball game where one person holds a stick and that person runs around a series of bases to come home if the person with the stick hits the ball well enough and hard enough, a lot of people try to catch it and prevent the batter from rounding the bases and get home. The president will soon announce a name for that and announce that he has invented this game. Also, he's invented rap music and the idea of taking a vacation in the summer if you're a school kid. I mean, I kid you not, the president proposed to reporters from the White House in all seriousness from the White House that he's thinking there should maybe be fireworks and a parade on the Fourth of July in Washington, D.C. It could catch on. It could become a tradition. & \multirow{2}{*}{\small 7.63} & \multirow{2}{*}{\small 0.10}\\   
\end{tabularx}
\caption{Snippets from MSNBC rated as highly uncivil by humans but with low toxicity score from Perspective. Human rating are between 1-10 and Perspective scores between 0-1.}
\label{tab:maddow_eg}
\end{table*}

\begin{table*}[t]
\centering
\scriptsize
\setlength{\tabcolsep}{4pt}
\begin{tabularx}{\linewidth}{L{1.0}|C{0.1}|C{0.1}}
\multicolumn{1}{c|}{\bf \small Snippets from FOX} & {\bf \small Human} & {\bf \small Perspective} \\ \hline
Meanwhile, this garbage deal that I've been telling you about, out of Washington, which allocates a measly additional \$1.375 billion for wall construction -- well, that has not been signed by the president. As a matter of fact, nobody in the White House has seen any language in the bill, none. So, let's be frank. It's not nearly enough money. Washington lawmakers, the swamp creatures, they have once again let we, the people, the American people down, this is a swamp compromise. Now, let me make predictions. Based on what the president is saying publicly, I love the press love to speculate, how did Hannity get his information? All right. We know this is a president that keeps his promises. And he goes at the speed of Trump. And if you watch every speech, if you actually listen to Donald Trump's words, you know, especially you people in the hate Trump media, you psychos, he telegraphs. He's saying exactly what he's planning to do. You're too stupid to listen. & \multirow{2}{*}{\small 8.13} & \multirow{2}{*}{\small 0.80} \\ 
\hline
The Democratic party with no plan to make your life better. More safe, more secure, more prosperous for not only you but for your children and you grandchildren. They are fueled too by a rage, hate and obsession with victimhood and identity politics, entitlement, pushing doom and gloom day and night, climate hysteria, indoctrinating your kids into believing the world is ending in 12 years. Just to push the socialist agenda. A party fueled by what can only be described as psychotic, literally unhinged, rage and hatred every second, minute and hour of the day against the duly elected president. They are not even trying to hide it. Just today we learned that shifty Adam Schiff, former MSNBC legal analyst, collusion conspiracy theorist Daniel Goldman to leave his new witch hunt against the President.  & \multirow{2}{*}{\small 9.00} & \multirow{2}{*}{\small 0.78} \\ 
\end{tabularx}
\caption{Examples of snippets from Fox, rated with human incivility ratings and Perspective.}
\label{tab:hannity_eg}
\end{table*}

There is only one snippet with Perspective toxicity score over 0.5 for the civil to borderline civil segments from the news shows; this indicates it has good precision for binary identification of civil content. Perspective's performance in detecting binary incivility (for snippets with ratings greater than 5) is mixed. The recall for incivility is not that good, with some of these snippets receiving low toxicity scores. The trend of increasing Perspective score with increasing human-rated incivility is observed mostly on segments from FOX. The incivility in FOX appears to be more similar to that seen in online forums, with name-calling and labeling of individuals. Some examples can be seen in Table \ref{tab:hannity_eg}. The more subtle, snarky comments from MSNBC are not detected as toxic by Perspective.

However, when characterizing shows by incivility, detecting utterances that may be problematic is not of much interest. The goal is to characterize the show (daily, weekly, monthly) overall. For this, we inspect the average incivility per show on the transcript and segment level (see Table \ref{tab:stats}). 
On both granularities, people perceive a statistically significant difference in civility between each pair of shows, with FOX perceived as the most uncivil, followed by MSNBC and PBS as the most civil. 

On the transcript level, the presence of incivility in PBS is not statistically significant. The raters chose 0.29 (fewer than one) uncivil utterances from PBS from all shows on the 17 days we study, compared with 4.5 per show for FOX. The 95\% confidence interval for the mean for uncivil utterances per show covers zero for PBS, so it is not statistically significant. The lower end of the  95\% confidence intervals of the mean transcript-level incivility in FOX and MSNBC is greater than zero, indicating consistent incivility in these programs. 

The segment-level analysis of civility ratings reveals the same trend, with a one-point difference between PBS and MSNBC and two points between MSNBC and FOX. All of these differences are statistically significant at the 0.01 level, according to a two-sided Mann-Whitney test.

The automated analyses paint a different picture. On the transcript level, FOX is overall the most uncivil, with about 6 speaker turns per predicted to be toxic, with a Perspective score greater than 0.5. PBS appears to be the next in levels of incivility, with more than one toxic turn per transcript. For both of these shows, incivility is over-predicted, and many of the segments predicted as uncivil are civil according to the human ratings. MSNBC is predicted to have fewer than one toxic turns per transcript, under-detecting incivility. On the segment level, FOX is again assessed as the most uncivil, and PBS again appears to be more uncivil than MSNBC. On the segment level, the differences are statistically significant. Perspective incorrectly characterizes PBS as significantly more uncivil than MSNBC.  

The correlation between human and Perspective incivility scores is 0.29 on the transcript and 0.51 on the segment level. Overall for broadcast news, Perspective cannot reproduce the incivility perception of people. In addition to the inability to detect sarcasm/snark, there seems to be a problem with over-prediction of the incivility in PBS and FOX. 

In the next section, we seek to establish some of the drivers of over-prediction errors, characterizing individual words as possible triggers of absolute or relative over-prediction of incivility. 

\begin{table}[t]
\centering
\small
\setlength{\tabcolsep}{4pt}
\begin{tabularx}{\linewidth}{L{0.1}R{0.15}|R{0.1}R{0.25}|R{0.1}R{0.25}}
\multirow{2}{*}{\bf Show} & \multirow{2}{*}{\bf Count} & \multicolumn{2}{c|}{\bf Human} & \multicolumn{2}{c}{\bf Perspective} \\
& & \textbf{Avg} & \textbf{95\% CI} & \textbf{Avg} & \textbf{95\% CI} \\ 
\hline
\multicolumn{6}{c}{\rule{0pt}{10pt} \bf Transcript Level} \\[1.5pt]
\hline
FOX & 17 & 4.53 & [2.62, 6.44] & 6.18 & [3.86, 8.49] \\
MSNBC & 17 & 1.24 & [0.40, 2.08] & 0.29 & [-0.01, 0.6] \\
PBS & 17 & 0.29 & [-0.06, 0.65] & 1.41 & [0.6, 2.23]\\

\hline
\multicolumn{6}{c}{\rule{0pt}{10pt} \bf Snippet Level} \\[1.5pt]
\hline
FOX & 116 & 7.09 & [6.85, 7.33] & 0.33 & [0.3, 0.37] \\
MSNBC & 44 & 4.97 & [4.43, 6.00] & 0.15 & [0.12, 0.17] \\
PBS & 59 & 3.87 & [3.56, 4.18] & 0.17 & [0.14, 0.19] \\ 
\end{tabularx}
\caption{Statistics for human incivility ratings and Perspective scores. Count is the number of transcripts and the number of snippets in each level of analysis.}
\label{tab:stats}
\end{table}

\section{Incivility Over-prediction}

Prior work has drawn attention to the fact that certain words describing people are incorrectly interpreted as triggers of incivility by Perspective, leading to errors in which a perfectly acceptable text segment containing the words would be predicted as toxic \cite{dixon2018measuring,hutchinson-etal-2020-social}. Their analysis, similar to other work on bias in word representations, starts with a small list of about 50 words to be analyzed in an attempt to find toxicity over-prediction triggers.

In our work, we seek to apply the same reasoning with the same goals, but in a more data-driven manner, without having to commit to a very small list of words for analysis. Given our text domain of interest (news) and the desiderata to characterize sources rather than individual text segments, we also find sub-errors, or words that do not lead to errors in toxicity prediction but have much higher than average toxicity associated with them compared to other words. 

Ideally, we would like to test the full vocabulary of a new domain for (sub-)error triggers, but methods for doing so do not exist and may not be practical when the vocabulary is too large. For this reason, we sample words in a way informed by the target application, choosing words that contribute the most to the average incivility score of one on the news sources compared to another of the sources. We sample a thousand words from each show. Occasionally, the same word is sampled for more than one show, so the final list for detailed analysis consists of 2,671 words.  

\subsection{Template-Derived Word Toxicity}

Then we proceed similarly to prior work, to fill short templates in which only one word differs, and a fixed slot is filled in with each word in turn. We use five templates, chosen in an ad-hoc way similar to prior work. We do not reuse templates from prior work because they are tailored for the analysis of identity words, while we would like any word from the vocabulary to be an acceptable completion of the template. The average toxicity of templates filled in with a given word now provides us with a word-specific incivility score by which we can compare the full set of selected words.

We fill in five short templates with each word in turn, to obtain a comparable incivility score for each word. The first two were chosen to be neutral; the third one contained a word with a possibly negative connotation but used in a neutral sense, the fourth contains negation, and the last one is a question, to account for a wider variety of possible contexts in which the incivility triggers may occur. The five templates we use are:

\vspace{+5pt}

{\em (i)} We wrote WORD on the page. 

{\em (ii)} They whispered WORD to themselves.  

{\em (iii)} I erased the word WORD from the page. 

{\em (iv)} You didn't write WORD. 

{\em (v)} Did he say WORD? 

\vspace{+5pt}
We use Perspective to predict the toxicity of each template when WORD is substituted with each of the words in our list for analysis. The average and maximum of these toxicities serve as a useful characterization of the words.

The templates were chosen so that the full sentence is not offensive, that is neither of the examples, even when filling it with an offensive word should be interpreted as deliberately rude or uncivil without any additional context. Higher values of predicted incivility would implicate the word substituted in the placeholder as the perceived trigger of incivility. We wish to find words where this association is wrong. 

\subsection{Offensive Words}

At the same time, without reference or knowledge of the template word-scores, we classified the words as offensive or not. Two of the authors classified words independently and then adjudicated the disagreements in a conversation. Ultimately, there were 65 words we judged as offensive out of the full list of 2,671 words. 

Separating offensive words from the rest is helpful. Using these words, we can get a sense of the degree to which Perspective incorporates context to make the prediction and then exclude them from further analysis. When templates are filled in with offensive words, the templates present hard cases for prediction. For these cases, context interpretation is necessary to make the correct prediction; word-spotting without incorporating context is likely to lead to an error. The ability to do this is important for our application: in the news, it is acceptable to report on someone else's incivility and their uncivil statements. If this is done with the purpose of expressing disapproval of the incivility, the reporting itself is not uncivil. 

Furthermore, the set of offensive words allows us to quantify the degree to which the template scores justify our choice of rules to classify non-offensive words as error and sub-error triggers. We will consider words to be error triggers if at least one template was judged by Perspective to have toxicity greater than 0.5. Sub-error triggers are words for which all five templates had toxicity lower than 0.5, but their average toxicity was markedly higher than that for other words. 

The average template toxicity for offensive words is 0.48, compared to 0.11 for the 2,606 non-offensive words in the list we analyzed. Of the 65 offensive words, 54\% had at least one template with toxicity greater than 0.5. Perspective clearly is not simply word-spotting to make the prediction. It produces toxicity scores below 0.5 for about half of the offensive words. For the other half, however, it often produces an error. 

In addition, 35\% of the offensive words met the criteria for sub-error trigger. Overall, 89\% of the offensive words meet either the error triggers or sub-error triggers criteria, confirming that these ranges of toxicity are the appropriate ones in which we should focus our attention in search of words that may have an unusually high association with toxicity. Example offensive words that met the sub-error criteria are: bozo, cheater, Crazy, crock, deplorables, F-ing, hoax, insane, mad, etc.

Other words, which we deemed non-offensive, have a template profile similar to that of the vast majority of offensive words. They are ones for which Perspective over-predicts toxicity. 

\begin{table*}
\centering
\scriptsize
\setlength{\tabcolsep}{4pt}
\begin{tabularx}{\linewidth}{L{1.0}}
\hline
0, 115-pound, abortion, abuse, abused, abusively, abysmal, accosted, adult, Africa, age-old, aliens, America, Americans, anti-semite, Anti-Trump, anti-Trump, Aryan, assaulted, assaulting, attack, baby, bad, barrier, Basically, beaten, belly, Black, black, Blow, blowing, bomber, bottom, bouncer, British, Brooks, brown, bunch, caliphate, capacity, cares, Catholic, Catholics, chicken, chief, child, children, China, Chinese, chock-full, Christian, church, Clinton, conforming, Content, COOPER, country, Covington, cows, crackers, crawling, creatures, cries, crime, crimes, criminal, CROSSTALK, cruelty, crying, DANIELS, dare, daughter, death, decrepit, defy, dehumanizing, Democrat, Democrats, demonize, denigrate, destroy-, died, Dingell, Dinkins, disrespectful, Donald, doomed, drug, drugs, Drunk, ducking, Duke, dumping, eggs, epidemic, European, evil, exist, explode, exploit, extremist-related, face, fake, Fallon, firearm, fithy, folks, Foreign, Former, FRANCIS, fraud, fry, gag, gagged, gang, gender, girls, governor, gun, guy, guy's, guys, handgun, harassed, harboring, hate, hate-, hate-Trump, hatred, head, heads, heartless, Hebrew, Hegel, her, herein, heroin, Hillary, HIV, horrors, hts, human, hush, Hymie, ifs, illness, imperialists, impugning, inaudible, inconclusive, infanticide, infuriating, inhumane, intelligence, interracial, invaders, Iran, Iraqi, ISIS, Islamophobic, Israel, Israelites, jail, Juanita, Kaine, Karine, kid, kids, Klan, Korea, laid, LAUGHTER, lie, lied, lies, life-and-death, life-death, limbs, litig, lying, MADDOW, MAGA-inspired, males, mama, man, men, mental, military, minors, missile, mock, mockery, molested, mouth, muck, N-, n't, NAACP, nation-state, needless, newscasters, Nonproliferation, nose, nuke, Obama, Obama's, obscene, obsessions, obsessive-compulsive, obsolete, organ, outrageous, ovations, Oversight, oxygen, p.m., painful, Pakistan, patriarchal, people, person, police, pope, President, president, president's, pretty, priest, priests, prison, prog, punched, punches, Putin, Putin's, queer, racial, Racism, racism, Rage, ranted, relations, religion, religious, relitigating, remove, REP., Republican, Republicans, RUSH, Russian, Russians, S, Saudi, savagely, self-confessed, self-defining, self-proclaimed, semitic, she, SHOW, sick, slavery, sleet, slurs, smear, socialist, son, Spartacus, stick, stop, stunning, supporters, supremacist, swamp, Syria, tampering, terror, terrorism, terrorists, thrash, threat, throat, tirade, toddler, TRANSCRIPT, trashing, treasonous, Trump, tumor, U.K., U.S, U.S., U.S.-backed, undress, unsuccessful, unvetted, upstate, vandalized, Vatican, Venezuela, Venezuelans, videotaped, videotaping, violated, violence, violent, VIRGINIA, virulent, voters, War, war, weird, welcome, Whitaker, White, white, WITH, woman, worse, worth, xenophobic, Yemen, you, your, yourself\\
\hline
\end{tabularx}
\caption{Sub-error trigger Words. The list comprises many identity words, words with negative connotations, words describing controversial topics, and words related to violence.}
\label{table:suberror_words}
\end{table*}

\section{Error Trigger Words for Perspective}

We consider a word to be an error trigger if at least one of the templates has a toxicity score of 0.5 or greater from Perspective. Below is the full list of error trigger words from the news shows. We informally organize them into categories to facilitate inspection.

\begin{small}
\begin{description}
\itemsep0em
\item[Identity] African-Americans, anti-Semites, anti-white, anti-women, BlacKkKlansman, feminism, Feminist, gay, Homophobia, Homophobic, homophobic, Islam, Islamic, Jew, LGBT, Muslim, women

\item[Violence and Sex] annihilation, assassinated, beheaded, die, kill, killed, killing, murdered, shooting; intercourse, pedophiles, pornography, prostitution, rape, raped, rapist, rapists, sex, sexist, Sexual, sexual, sexually, sodomized

\item[Ambiguity] Dick, dirty, garbage, rats

\item[Informal] dopest, farting

\item[Other] brain, counter-intelligence, blackface, hypocrisy

\end{description}
\end{small}

Many error triggers are identity words describing people. Many of these identities, like gay and Muslim, were known triggers of incorrect toxicity prediction, and Perspective was specially altered to address this problem \cite{DBLP:conf/aies/DixonLSTV18}. Clearly, however, they are still a source of error, and the approaches for bias mitigation have not been fully successful. As we mentioned in the introduction, a system that is unstable in terms of its predictions when identity words appear in a text is not suitable for analysis of political discourse, where identities are mentioned often and where in many cases, it is of interest to quantify the civility with which people talk about different identity groups. 

The second large class of error triggers consists of words related to death and sex. In online comments, toxic statements are often threats of sexual violence and death. Perhaps this is why words broadly related semantically are associated with toxicity. These words, however, can appear in news contexts without any incivility. Reports of violence at home and abroad are a mainstay of news, as well as reports of accusations of sexual misconduct and abuse. A system that is not capable of distinguishing the context of usage is not going to provide reliable quantification of incivility in the news.    

Similarly, the ambiguous error triggers are words with more than one meaning, which could be offensive but can be used in civil discussion as well. For these, Perspective has to disambiguate based on the context. All of the templates we used were neutral, so for these, Perspective makes an error. For example, the name `Dick' is an error trigger. The word indeed has an offensive meaning, but in the news shows we analyze, all mentions are in the sense of a person's name.

A couple of error-triggers are informal, clashing more in register than conveying incivility.  

\section{Sub-Error Triggers}

Sub-error triggers of incivility are words that are not offensive and for which Perspective returns a toxicity score below 0.5 for each of the five templates when the word is plugged in the template. The error triggers we discussed above lead to actual errors by Perspective for its intended use. The sub-error triggers are in the acceptable level of noise for the original purpose of Perspective but may introduce noise when the goal is to quantify the typical overall civility of a source.

To establish a reference point for the expected average template civility of non-offensive words, we sample 742 words that appeared in at least 10 speaker turns (i.e., were fairly common) and appeared in speaker turns in the news shows that received low toxicity predictions from  Perspective, below 0.15. These were part of the list we classify as offensive or not. No word in this reference sample was labeled as offensive.

The average template score for the reference sample is 0.09, with a standard deviation of 0.01. We define sub-error triggers to be those whose average template toxicity score is two standard deviations higher than the average in the reference list.  There are 325 non-offensive words that meet the criteria for sub-error triggers. They are shown in Table \ref{table:suberror_words}. The list is long, which is disconcerting because it is likely that sentences containing multiple of these words will end up triggering errors.   

\begin{table*}[t]
\centering
\scriptsize
\setlength{\tabcolsep}{4pt}
\begin{tabularx}{\linewidth}{L{1.0}}
The other thing that's important, both sides seem to be mad about this. On the conservative side, you have conservative voices who are saying, this deal is \textbf{pathetic}, it's an insult to the president. On the Democratic side and on the liberal side, I have activists that are texting me saying, Nancy Pelosi said she wasn't going to give \$1 for this wall, and now she is.\\
\hline
I do feel safe at school. And I know that sounds kind of \textbf{ridiculous}, since tomorrow makes a year since there was a shooting at my school. But I do feel safe at school. And I feel safe sending my children to school. I know that there are recommendations that have been made to arm teachers, and I think that is the \textbf{stupidest} thing that I have ever heard in my life. Me having a gun in my classroom wouldn't have helped me that day.\\
\hline
VIRGINIA ROBERTS, Victim: It ended with sexual abuse and intercourse, and then a pat on the back, you have done a really good job, like, you know, thank you very much, and here's \$200. You know, before you know it, I'm being lent out to politicians and to academics and to people that -- royalty and people that you just -- you would never think, like, how did you get into that position of power in the first place, if you're this \textbf{disgusting}, evil, decrepit person on the inside?\\
\hline
In a major incident like this, at first there's, you know, there's sort of a stunned numb thing that happens. And then you kind of go into this honeymoon phase. There's just a high level of gratefulness for all of the help that's coming. And then you get to the phase that we're kind of beginning to dip in now, which is life \textbf{sucks} right now, and I don't know how long it's going to \textbf{suck}.\\
\hline
It's \textbf{insane}. We are throwing away tremendous amounts of food every day. And there are people next door a block away that aren't getting enough food.\\
\end{tabularx}
\caption{Segments from PBS that contain offensive words (marked in boldface).}
\label{tab:pbs_eg}
\end{table*}

A sizeable percentage of sub-error triggers are related to identities of people---gender, age, religion, country of origin. Oddly, a number of child-describing words appear in the list (baby, child, kid, toddler). There are also several personal names in the list, which indicates spurious correlations learned in Perspective; names by themselves can not be uncivil. 

Second person pronouns (you, your) and third-person feminine pronouns (her, she) are sub-error triggers. The second-person pronouns are likely spuriously associated with toxicity due to over-representation in direct toxic comments directed to other participants in the conversation. Similarly, the association of female pronouns with toxicity is likely due to the fact that a large fraction of the indirect toxic comments online are targeted to women.

Regardless of the reasons why these words were associated with incivility by Perspective, the vast majority of them are typical words mentioned in the news and political discussions, explaining to an extent why Perspective is not able to provide a meaningful characterization of civility in the news. 

\begin{table}[t]
\centering 
\footnotesize
\setlength{\tabcolsep}{2pt}
\begin{tabularx}{\linewidth}{L{0.18}|R{0.2}R{0.2}R{0.2}|R{0.1}}
\textbf{Category} & \multicolumn{1}{c}{\bf FOX} & \multicolumn{1}{c}{\bf MSNBC} & \multicolumn{1}{c|}{\bf PBS} & {\bf Total} \\ 
\hline
Error            & 197 {[}.44{]}  & 55 {[}.12{]}  & 196 {[}.44{]}  & 448                                \\
Sub-error        & 1537 {[}.39{]} & 723 {[}.18{]} & 1708 {[}.43{]} & 3968                               \\
Offensive        & 277 {[}.52{]}  & 101 {[}.19{]} & 156 {[}.29{]}  & 534                               
\end{tabularx}
\caption{Number [and fraction] of segments containing at least one (sub-)error trigger or offensive word.}
\label{tab:word_distributions}
\end{table}

Table \ref{tab:word_distributions} shows the distribution of error triggers, sub-error triggers, and offensive words in the three programs. Most of the segments containing error and sub-error triggers are in FOX and PBS; this could explain the observation that incivility is much higher in FOX compared to the other two programs when analyzed with Perspective than compared to that from human judgments. This also explains why PBS, judged significantly more civil than MSNBC by people, appears to be somewhat less civil. Not only is Perspective not able to detect some of the incivility in sarcasm present in MSNBC, but also PBS segments include substantially more (sub-)error triggers than MSNBC.

More segments from PBS contain uncivil words, compared to MSNBC. Table \ref{tab:pbs_eg} shows some representative PBS segments with offensive words. They are often reported incivility, or occur in segments where a guest in the program uses such language. Most of the segments are not overall uncivil.

\section{Conclusion}

The work we presented was motivated by the desire to apply off-the-shelf methods for toxicity prediction to analyse civility in American news. These methods were developed to detect rude, disrespectful, or unreasonable comment that is likely to make you leave the discussion in an online forum. To validate the use of Perspective to quantify incivility in the news, we create a new corpus of perceived incivility in the news. On this corpus, we compare human ratings and Perspective predictions. We find that Perspective is not appropriate for such an application, providing misleading conclusions for sources that are mostly civil but for which people perceive a significant overall difference, for example, because one uses sarcasm to express incivility.  Perspective is able to detect less subtle differences in levels of incivility, but in a large-scale analysis that relies on Perspective exclusively, it will be impossible to know which differences would reflect human perception and which would not.  

We find that Perspective's inability to differentiate levels of incivility is partly due to the spurious correlations it has formed between certain non-offensive words and incivility. Many of these words are identity-related. Our work will facilitate future research efforts on debiasing of automated predictions. These methods start off with a list of words that the system has to unlearn as associated with a given outcome. In prior work, the lists of words to debias came from informal experimentation with predictions from Perspective. Our work provides a mechanism to create a data-driven list that requires some but little human intervention. It can discover broader classes of bias than people performing ad-hoc experiments can come up with. 

A considerable portion of content marked as uncivil by people is not detected as unusual by Perspective. Sarcasm and high-brow register in the delivery of the uncivil language are at play here and will require the development of new systems.

Computational social scientists are well-advised to not use Perspective for studies of incivility in political discourse because it has clear deficiencies for such application. 

\section*{Acknowledgments}
We thank the reviewers for their constructively critical comments. 
This work was partially funded by the National Institute for Civil Discourse. 

\bibliography{anthology,eacl2021}
\bibliographystyle{acl_natbib}

\end{document}